\documentclass[11pt]{article}

\usepackage[final]{ACL2023}

\usepackage{times}
\usepackage{latexsym}

\usepackage[T1]{fontenc}
\usepackage[utf8]{inputenc}

\usepackage{microtype}

\usepackage{inconsolata}
\usepackage[table,xcdraw]{xcolor}

\usepackage{algpseudocode}
\usepackage{algorithm}

\usepackage{graphicx}
\usepackage{amsmath}
\usepackage{mathtools, amssymb, lipsum, multicol}
\usepackage{soul}
\usepackage{todonotes}
\usepackage{array}
\usepackage{hyperref}
\usepackage{multirow}
\usepackage{makecell}
\usepackage{scalerel}

\usepackage{float} 
\usepackage{fontawesome5}
\newlength{\myMheight}
\makeatletter
\let\old@footnotemark\@footnotemark
\renewcommand{\@footnotemark}{\hbox{\githubmarker}\old@footnotemark}
\makeatother

\DeclareMathOperator{\enc}{\mathbf{E}} % Define the encoder operator

\title{Triple-Encoders: Representations That Fire Together, Wire Together}

\author{Justus-Jonas Erker$^{1\;2}$,  Florian Mai$^{3}$, Nils Reimers$^{4}$, \\
     {\bf Gerasimos Spanakis}$^{2}$, {\bf Iryna Gurevych}$^{1}$
     \\ 
     $^{1}$Ubiquitous Knowledge Processing Lab (UKP Lab) \\
     Department of Computer Science and Hessian Center for AI (hessian.AI) \\
     Technical University of Darmstadt \\
     $^{2}$Maastricht University, $^{3}$KU Leuven, \\
     $^{4}$Cohere \\
    \url{www.ukp.tu-darmstadt.de} 
     }

\newcommand{\githublogo}{\protect\raisebox{-1pt}{\includegraphics[height=0.35cm]{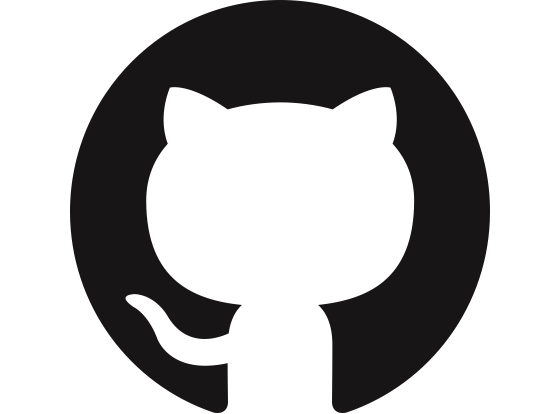}}}
\newcommand{\huggingfacelogo}{\protect\raisebox{-1pt}{\includegraphics[height=0.32cm]{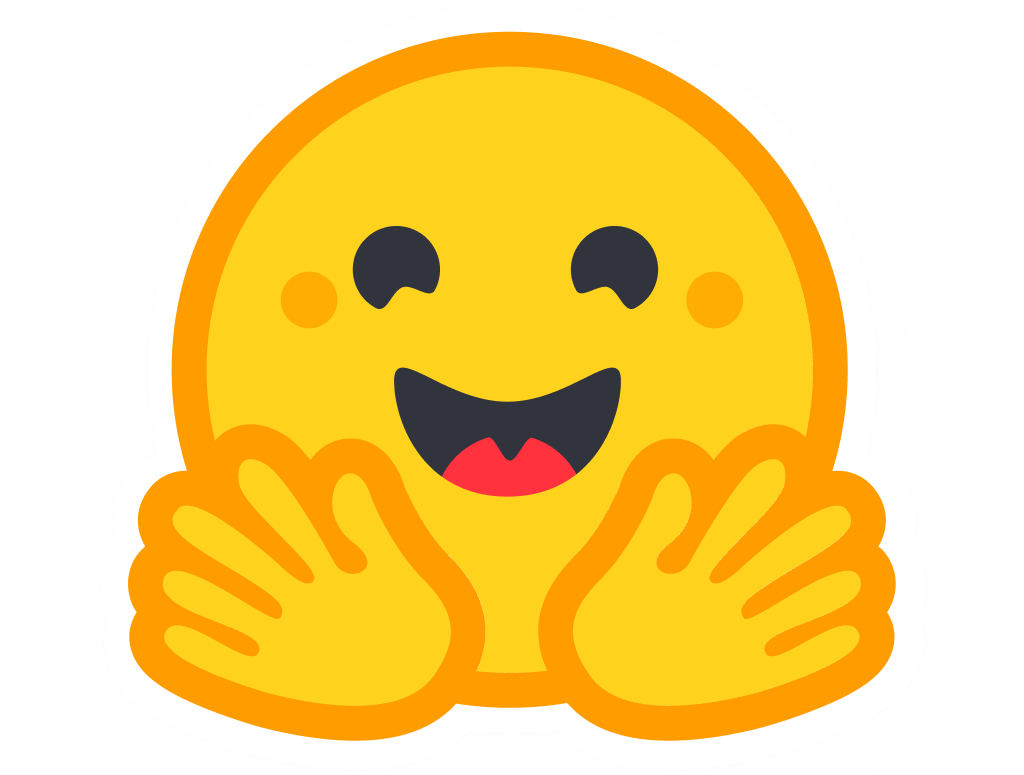}}}

\begin{document}
\maketitle
\hspace*{\fill} \\
\hspace*{\fill} \\
\begin{abstract}
\iffalse
\todo{The abstract is very, very difficult to understand if it's the first thing a reviewer is reading. I think it needs to be rewritten. Try to employ the same writing strategy as for the introduction, just with less details!}
This study introduces the innovative triple-encoders approach to conversational sequence modeling, a method that surmounts the inherent limitations of conventional models like ConveRT, particularly their weak latent space interaction and repetitive context recomputation. Rooted in Curved Contrastive Learning (CCL), triple-encoders revolutionize dialog representation by dividing the context into distinct sub-latent spaces and employing a Hebbian-inspired co-occurrence learning mechanism. This innovative strategy enables the independent encoding of utterances into a latent space and the composition of sequences for retrieval, remarkably without any additional weights. Employing straightforward yet potent operations: mean pooling, batch matrix multiplication for computing similarity, and summing across the sequential dimension. triple-encoders demonstrate a significant improvement in sequence modeling. Specifically, they exhibit a 36\% enhancement in open-dialog and a 46\% improvement in task-oriented scenarios in terms of average rank, compared to previous CCL models. Furthermore, the model's robust architecture facilitates in better planning performance and  generalization to zero-shot settings, showcasing its versatility and broad applicability. 
\fi

Search-based dialog models typically re-encode the dialog history at every turn, incurring high cost.
Curved Contrastive Learning, a representation learning method that encodes relative distances between utterances into the embedding space via a bi-encoder, has recently shown promising results for dialog modeling at far superior efficiency.
While high efficiency is achieved through independently encoding utterances, this ignores the importance of contextualization. 
To overcome this issue, this study introduces triple-encoders, which efficiently compute distributed utterance mixtures from these independently encoded utterances through a novel hebbian inspired co-occurrence learning objective in a self-organizing manner, without using any weights, i.e., merely through local interactions. Empirically, we find that triple-encoders lead to a substantial improvement over bi-encoders, and even to better zero-shot generalization than single-vector representation models without requiring re-encoding. Our code\footnotemark[1] and model\footnotemark[2] are publicly available.

\footnotetext[1]{\href{https://github.com/UKPLab/acl2024-triple-encoders}{\githublogo\hspace{0.1cm}UKPLab/Triple-Encoders}}
\footnotetext[2]{\href{https://huggingface.co/UKPLab/triple-encoders-dailydialog}{\huggingfacelogo\hspace{0.05cm} UKPLab/Triple-Encoders-DailyDialog}}

\end{abstract}
\section{Introduction}

\begin{figure}[t]
    \centering
    \hspace*{\fill} \\
    \hspace*{\fill} \\
    \includegraphics[width=0.5\textwidth]{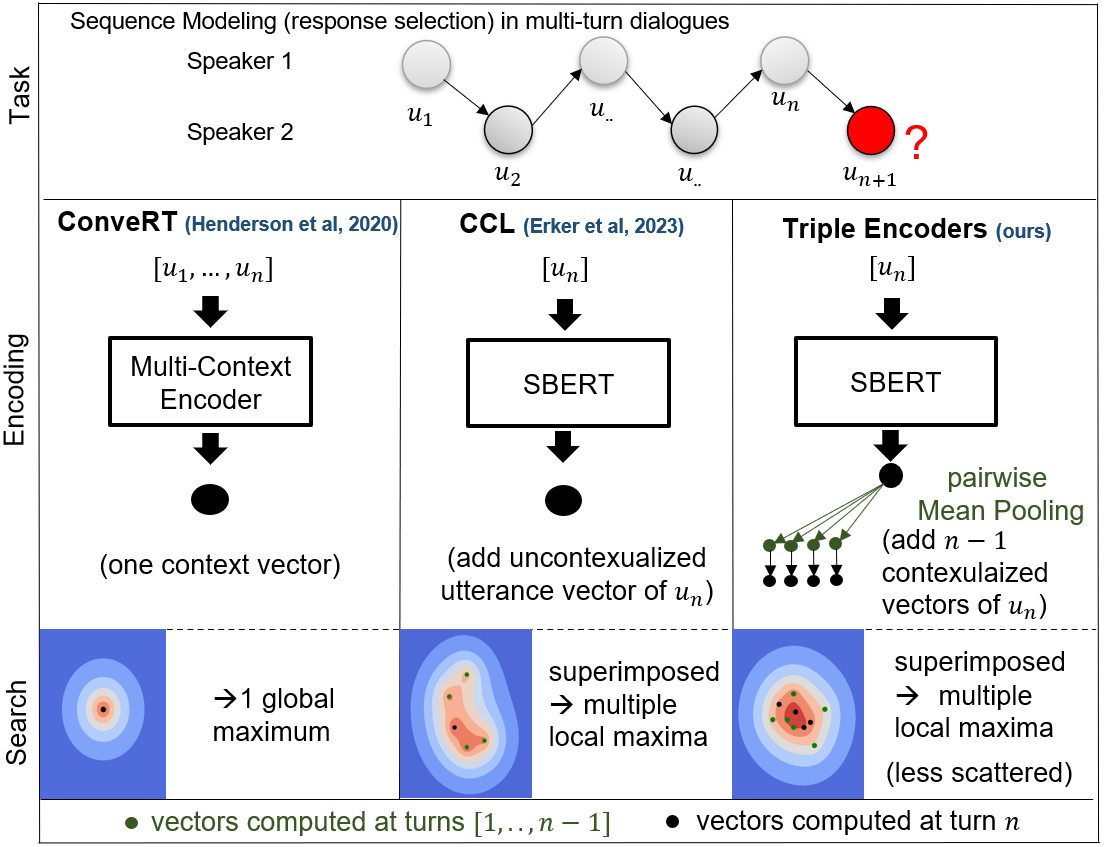}
    \caption{Comparison of our Triple Encoder to \citet{henderson-etal-2020-convert} and \citet{erker-etal-2023-imagination}. Similar to CCL we only need to encode and compute similarity scores of the latest utterance. At the same time, we achieve contextualization through pairwise mean-pooling with previous encoded utterances combining the advantages of both previous works. Our analysis shows that the co-occurrence training pushes representations that occur (\textbf{fire}) together closer together, leading to stronger additive properties (\textbf{wiring}) when being superimposed (compared to \citet{erker-etal-2023-imagination}) and thus to a better next utterance selection.}
    \label{fig:intro}
\end{figure}

Traditional search-based approaches in conversational sequence modeling like ConveRT \citep{henderson-etal-2020-convert} represent the entire context (query) in one context vector (see Figure \ref{fig:intro}). 
This has two major drawbacks:
\textbf{(a)} Recomputing the entire vector at each turn is computationally expensive, and
\textbf{(b)} it is difficult to compress the context's relevant information for any possible candidate response into a single vector. Furthermore, the encoder models are limited to a maximum number of tokens, usually 512.

\begin{figure}[h]
    \centering
    \includegraphics[width=0.4\textwidth]{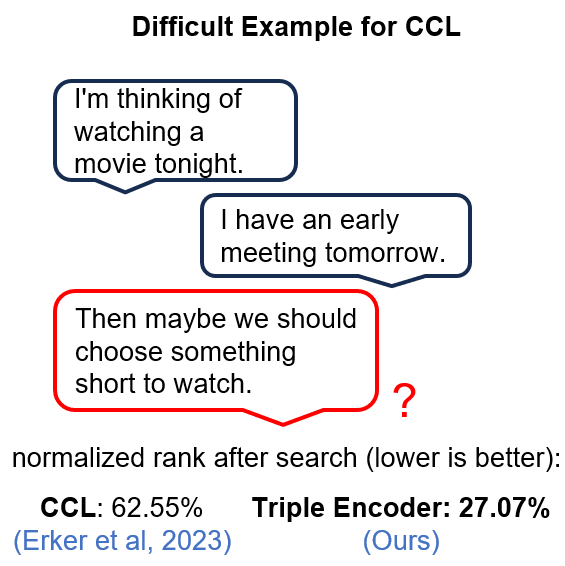}
    \caption{ Difficult example for next utterance selection based on solely independent utterances. Here the model must know that both utterances occur together as it requires considering them jointly to derive the third utterance (in red). This is reflected by the significant gap in the normalized rank between our contextualized approach and the uncontextualized approach of \citet{erker-etal-2023-imagination}.}
    \label{fig:difficult}
    %\todo{@IG this is not new, just moved from the appendix}
\end{figure}

Curved Contrastive Learning (CCL) \citep{erker-etal-2023-imagination} demonstrated that it is possible to encode utterances separately in a latent space and accumulate sequence likelihood based on solely cosine similarity, thanks to treating cosine similarity not as a semantic but as a directional relative dialog turn distance measure between utterance pairs (through two sub-spaces representing a temporal direction: \textbf{before} and \textbf{after}). This relativistic approach tackles \textbf{(a)}, by enabling sequential search with a constant complexity, as \textbf{only} the latest utterance needs to be encoded and computed during inference as shown in Figure \ref{fig:intro}. \textbf{(b)} Furthermore, each candidate utterance can interact with every independently projected utterance, allowing a richer interaction. However, encoding utterances independently means they are not contextualized, disregarding a crucial feature of conversation. An example is illustrated in Figure \ref{fig:difficult}.

For the first time, in this paper we propose a method that contextualizes utterance embeddings in dialog sequences in a self-organizing manner, \emph{without the use of additional weights}, i.e, merely through local interactions (in form of efficient vector algebra) between separately encoded utterances after appropriate pre-training.
While previous work has shown that mean pooling is a strong method for sentence composition from tokens~\citep{pagliardini-etal-2018-unsupervised,reimers-gurevych-2019-sentence}, we demonstrate that this can be generalized to a higher abstraction level: distributed pairwise sequential composition (illustrated in Figure \ref{fig:intro}). To realize this, we present triple-encoders, which segment the context space of CCL into two distinct latent spaces denoting the relative order of utterances in the context. By linearly combining (averaging) representations from these sub-spaces through a co-occurrence learning objective, we create new contextualized embeddings that we can incorporate into CCL, resulting in Contextualized Curved Contrastive Learning (C3L). At inference time, our method efficiently contextualizes independently encoded utterances based on solely local interactions (\textbf{without any additional weights}): Our method applies only \textbf{(1) mean pooling}, a \textbf{(2)} \textbf{matrix multiplication} for computing the similarity and one \textbf{(3) summation} (across the sequential dimension) operation to aggregate similarity scores.

While we focus on modeling dialog in this paper, the sequential modularity of our method can in principle be used for any text sequence. Pilot experiments on next sentence selection of children stories are reported in Appendix \ref{sec:CBT}, while we leave thorough exploration to future work. Our experiments are aimed at the following research questions:

\textbf{RQ1:} What is the effect of triple-encoder training (C3L) + triple encoder at inference compared to CCL?

\textbf{RQ2:} What is the effect of triple-encoder training (C3L) while encoding utterances at inference time without contextualization (like CCL)? % while I.e. using the triple encoder as bi-encoder (CCL) during inference. 

Our experimental results suggest that our approach improves substantially over standard CCL. % while generalizing to task-oriented dialogues. 
Notably, our method outperforms ConveRT \citep{henderson-etal-2020-convert} in a zero-shot setting while our method requires no additional learnable parameters for contextualization. While triple-encoder training alone improves the performance considerably (RQ2), using triple-encoder contextualization at inference time (RQ1) leads to additional performance gains while keeping linear complexity.

%%% WE DON'T NEED THE FOLLOWING PARAGRAPH IMO
%Furthermore, we demonstrate how we can reduce the index size during inference through pre-filtering, to our best knowledge a previously unexplored task.
%\todo{finished until here. Todo: structure}
%We structure the paper as follows, in section \ref{sec:} we discuss the related work. The methodology, including the evaluation, baselines is discussed in section §\ref{sec:method}. We discuss the architecture and the learning objective of triple-encoders in section §\
\section{Related Work} \label{sec:related}
We will start with related work on embedding compositionality, conversational sequence modeling and self-organizing maps. Next, we will describe retrieval methods that have been a motivation to our distributed representations. Lastly, we will discuss CCL as the core foundation of our work. 
\subsection{Composition and Self-Organization}
% start with weight-less compositionality
%For the first time, our method contextualizes utterance embeddings in dialog sequences \emph{without the use of additional weights}, i.e, merely through the use of efficient vector algebra (local interactions) after appropriate pre-training.
% there is a lot of research on compositionality for word embeddings, etc
Weight-less compositionality of embeddings is a well-studied problem for word representations~\citep{mitchell-lapata-2008-vector, rudolph-giesbrecht-2010-compositional,mikolov2013distributed, mai2018cbow}, but has received little attention for larger text units such as sentences or utterances.
For these, investigations are limited to small contexts such as pairwise sentence relations (e.g. NLI)~\citep{sileo-etal-2019-composition} or sentence fusion~\citep{huang-etal-2023-bridging}, and are outperformed by parameterized composition operators. In the context of conversational sequence modeling, conventional methodologies typically employ parameterized functions (learned weights) that act as an external force to contextualize utterance embeddings that are computed independently~\citep{liu-etal-2022-unsupervised, zhang-etal-2022-history}. As far as we are aware, this is the first method in which contextualization in conversational sequence modeling has been achieved solely through local interactions, without the reliance on additional weights, where all information is stored in the geometry of the latent space. This approach aligns with the self-organization principle found in nature that 
\begin{quote}
     describes the emergence of global order from local interactions between components of a system without supervision by external directing forces \cite{RezaeiLotfi2019}
\end{quote}
demonstrating how global order within dialogue sequences can emerge from localized interactions (mean pooling and cosine similarity) among utterance embeddings. The self-organization principle has previously been applied to machine learning in self-organizing maps \cite{kohonen1982self}.

%- weight based approaches matrix --> contribution first entirely non-weighted approach
%\url{https://aclanthology.org/J19-1005.pdf}
%- \citep{liu-etal-2022-unsupervised} --> hierachical transformer (1 transformer layer contextualization) 
%- similar \url{https://aclanthology.org/2022.findings-emnlp.247.pdf} \citep{zhang-etal-2022-history} creates history memory matrix that is added to a the current turn encoder but uses additional con
%While previous work on Hierarchical Transformers in conversation modeling have shown that turn-wise representations enable significant contextualization and efficiency gains \citep{liu-etal-2022-unsupervised} \citep{zhang-etal-2022-history}, all these techniques require complex self-attention architectures on top making them unsuitable for large scale search.

\subsection{Retrieval}
%Convert has weak interaction (through weak interaction)
Typically, neural response retrieval systems like ConveRT~\citep{henderson-etal-2020-convert} (see Figure~\ref{fig:intro}) produce a single context embedding per turn that is then compared to candidate utterance embeddings.
This leads to weak interactions with candidate utterances as not all information can be compressed into one vector.
Previous work in retrieval has addressed the weak interaction of bi-encodings through several techniques. Previous work like MORES \citep{gao-etal-2020-modularized}, PreTTR \citep{10.1145/3397271.3401093} or PolyEncoders \citep{humeau2020polyencoders} tackled this problem by encoding each candidate representation with a query via a late-stage self-attention mechanism to enable a richer interaction. Though this technique outperformed traditional bi-encoders, the attention mechanism does not scale with large search spaces. Another technique that was the inspiration for our average and maximum similarity based approaches is ColBERT \citep{10.1145/3397271.3401075} and ColBERTV2 \citep{santhanam-etal-2022-colbertv2} which has shown that this concept works well on word token level. %When it comes to conversational based approaches, many 
%\subsection{Dialog Modelling Planning}

%\subsection{Hebbian Learning}
%Our work highlights the potential of harnessing the fire-wire principle of Hebb \citep{https://doi.org/10.1002/sce.37303405110}, previously adapted as an alternative to gradient descent \citep{melchior2019hebbiandescent} or in continual representation learning \citep{morawiecki2022hebbian} can now be seen influencing the self-organization of sequential properties in representation learning.
%\section{Curved Contrastive Learning} \label{sec:background}
%In this section, we first review the previous work on Curved Contrastive Learning (CCL). We then explain how our triple-encoder addresses the shortcomings of CCL.

\subsection{Uncontextualized CCL  via Bi-Encoder}
\begin{figure}[h]
    \centering
    \includegraphics[width=0.5\textwidth]{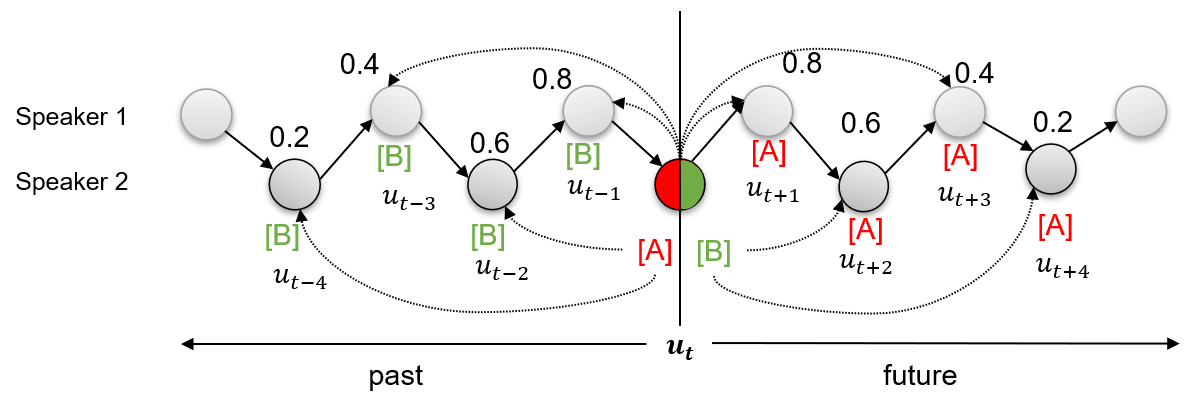}
    \caption{Concept of relativity in Imaginary Embeddings with $w=5$ using \textbf{before} \texttt{[B]} and \textbf{after} tokens [A] \citep{erker-etal-2023-imagination}}
    \label{fig:CCLfuturepast}
\end{figure}
We build upon the previous work on Curved Contrastive Learning (CCL)~\citep{erker-etal-2023-imagination}, a self-supervised representation learning technique based on sentence embedding methods like SentenceBERT \citep{reimers-gurevych-2019-sentence}. Similar to how our universe is made up of a stage between space and time, CCL learns a stage between semantics and the directional relative turn distance of utterance pairs in multi-turn dialog. 
As Figure~\ref{fig:CCLfuturepast} illustrates, the resulting embeddings are inspired by the concept of relativity \citep{Einstein1921-EINRTS-2}: By embedding utterances with special \textbf{before} (\texttt{[B]}) and \textbf{after} (\texttt{[A]}) tokens into two distinct subspaces, directional temporal distances become relative to the observer. Concretely, as Figure \ref{fig:CCLfuturepast} shows, when traveling through this space from $t-1$ to the next turn $t$, CCL linearly decreases the similarity to every previous utterance $u_s, s<t$ and increase the similarity to every $u_s, s>t$ as part of the sequence.
Formally, given a sequence of utterances $u_0, u_1, ... ,u_n$, choose a window size $w$, then the pretraining objective of CCL is
\begin{equation*}
 cos(\enc([B] \ u_i), \enc([A] \ u_k)) = 1-\frac{k-i}{w},
\end{equation*}
which we enforce through an MSE loss for $0 < k - i < w$. $\enc$ refers to a text encoder such as SBERT. We also refer to this model as the bi-encoder as it uses a dual encoder.
This training objective together with directional and random hard negatives shows strong performance in sequence modeling and planning tasks~\citep{erker-etal-2023-imagination}.
While ConveRT encodes $t$ utterances at step $t$, resulting in an overall complexity of $\mathcal{O}(n^2)$ utterance encodings, this approach only encodes one new utterance at every step, resulting in an overall complexity of $\mathcal{O}(n)$ utterance encodings.

However, we hypothesize that the lack of contextualization does not reflect the dialogs' highly contextual dependency which prohibits even better performance. With our triple-encoder we address this core limitation as we will show empirically in this paper.

%\todo{We need to again point out what the problem with CCL is!! Repeat the problem!111}

\section{Contextualized CCL via Triple-encoders}\label{sec:triple}
\begin{figure}[h]
    \centering
    \includegraphics[width=0.5\textwidth]{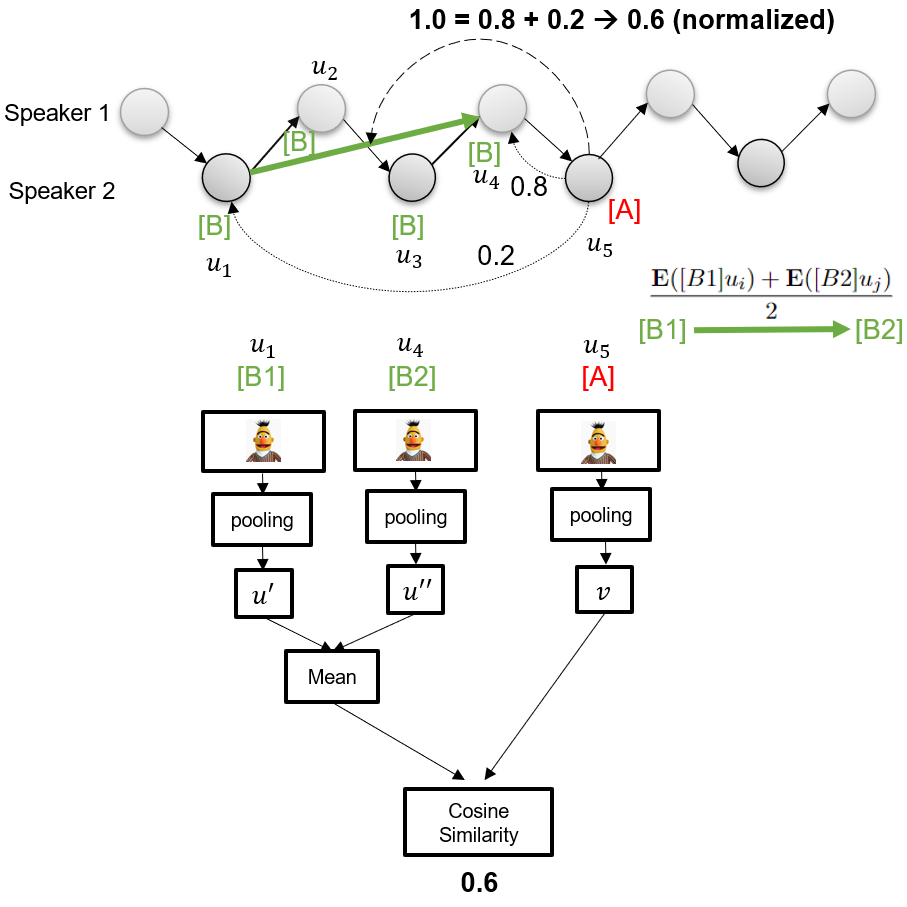}
    \caption{Our Triple-Encoder architecture with two directional \textbf{before} tokens \texttt{[B1]} and \texttt{[B2]}. We create a combined state of two utterances as the average between the separately encoded embeddings. The target distance of this new combined state results as a normalized sum of each individual utterance score from the bi-encoder Curved Contrastive Learning.}
    \label{fig:triple_arch}
\end{figure}
\noindent
We design an extension to the CCL framework that addresses the aforementioned issue while retaining the same order of encoding complexity at inference as the bi-encoder model ($\mathcal{O}(n)$), resulting in Contextualized Curved Contrastive Learning (C3L).

To enhance the CCL embeddings with contextualization, two additional special tokens are added to the \textbf{before} space, \texttt{[B1]} and \texttt{[B2]}. These tokens denote the relative order of the utterances in the dialog, i.e. $[B1] u_i \wedge [B2] u_j \Leftrightarrow  i<j$ to add positional information. To model the interaction between these different context utterances, we choose a simple mean operation as shown on the right of Figure~\ref{fig:triple_arch}. Given that the \texttt{[B1]} and \texttt{[B2]} tokens create distinct representations, they're effectively projecting the utterance into two different embedding spaces. By combining utterance states of \texttt{[B1]} and \texttt{[B2]} through a mean operation, a new combined state that carries information from both original states is created.

\begin{figure}[h]
    \centering
    \includegraphics[width=0.5\textwidth]{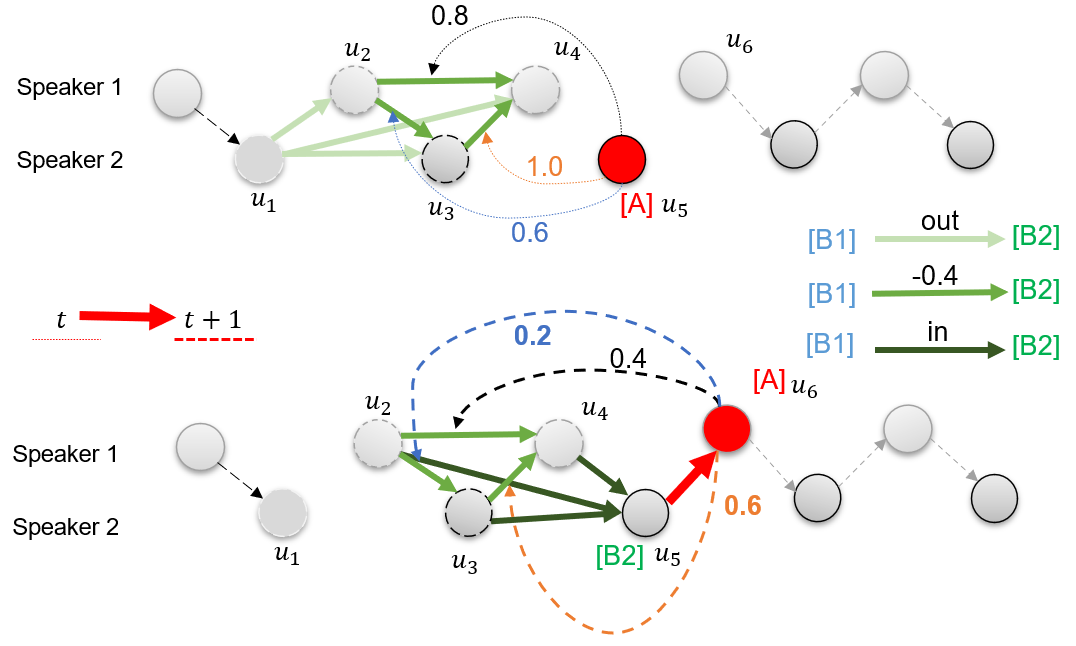}
    \caption{Relative time dimension in our proposed Contextual Curved Contrastive Learning. As the observation window moves from $t \rightarrow t+1$, $3$ new triplets are added (dark green), $3$ removed (light green), and $3$ decayed by $-0.4$ (green). As shown through the incoming green arrows at utterance $u_5$, we only have to encode the new incoming utterance with a \texttt{[B2]} token. In the next turn we require the \texttt{[B1]} token that can be encoded at idle time while the dialog partner is speaking.}
    \label{fig:relativeTimeDimension}
\end{figure}

\citet{erker-etal-2023-imagination} have shown that the curved scores (a linear decrease of similarity) outperform classic hard positives/negatives in sequence modeling and enhance sequential information that enables better planning capability. To keep the phenomenon of moving in a relativistic fashion through a continuous temporal dimension with respect to the \textbf{after} space, we construct the pairwise utterance mixture distance to a following utterance as the average of the two individual distances, as defined in CCL. Formally, for window size $w$ and $0 < i < j < k$ and $k - i < w$, if the distance between $[B1]u_i$ and $[A]u_k$ is $1-\frac{k-i}{w}$, and the distance between $[B2]u_j$ and $[A]u_k$ is $1-\frac{k-j}{w}$, then the joint representation of utterances in the \textbf{before} space should have distance $norm(2-\frac{2k-(i+j)}{w})$. Hence, we enforce for positive examples:
\begin{align}\label{eq:train}
\begin{split}
    & \cos\left(\frac{\enc([B1] u_i) + \enc([B2] u_j)}{2}, \enc([A] u_k)\right) \\
    = & \operatorname{norm}\left(2-\frac{2k - (i+j)}{w}\right),
    \end{split}
\end{align}
where $\operatorname{norm}$ normalizes the values from $[\frac{1}{w}, 2- \frac{3}{w}]$ to $[\frac{1}{w},1]$ via min-max scaling to match the range of cosine similarity. Figure~\ref{fig:triple_arch} illustrates this procedure.
Like in CCL, this objective can be computed efficiently when moving from step $t-1$ to $t$, since only the last utterance has to be encoded at every inference step, as Figure~\ref{fig:relativeTimeDimension} illustrates.

\subsection{Co-occurrence Learning Through Hard Negatives}
With the positive training examples from above, the model does not necessarily have to learn co-occurrence information, because it suffices to identify one context utterance in the input to reach a low training error. We introduce hard negative examples to mitigate this. By training every true context representation (both \texttt{[B1]} and \texttt{[B2]}) with one random utterance as hard negatives, we enable a novel co-occurrence learning paradigm that only lets a candidate representation (in the after space) \textbf{wire} to its mixed contextualized representation if \textbf{both} context representations \textbf{fire} in a sequence \textbf{together}. 
Hard negatives are constructed from random utterances $u_r, u_{r}'$ sampled from the training set:
%- write about the co-occurence learning objective 
\begin{equation}
\begin{array}{lcl}
\cos\left(\frac{\enc([B1] u_i) + \enc([B2] u_r)}{2}, \enc([A] u_k)\right) & = & 0.0 \\
\cos\left(\frac{\enc([B1] u_r) + \enc([B2] u_i)}{2}, \enc([A] u_k)\right) & = & 0.0 \\
\cos\left(\frac{\enc([B1] u_r) + \enc([B2] u_r')}{2}, \enc([A] u_k)\right) & = & 0.0  \\
\end{array}
\end{equation}
These are generated for every $1 \leq i < k$ and $k-i<w$. 
%This way we construct 6 positives to 12 negatives (with a ratio of $\frac{1}{3}$:$\frac{2}{3}$). 

% additional losses
\iffalse
The triple-encoder is furthermore pre-trained as bi-encoder using the curved contrastive learning objective \citep{erker-etal-2023-imagination} before training it on the triplets. This way the model is trained with additional directional negatives in the pre-training stage. However, we will also experiment with versions that a trained from scratch as well as include directional negatives in the triplet training. Alike curved contrastive learning, the NLI learning objective is added as a multi-training objective during fine-tuning. As \citet{erker-etal-2023-imagination} has shown that speaker tokens in the before space for indicating odd \texttt{[O]} and even \texttt{[E]} turn distances to the utterance in the after space are improving sequence modeling, we train all triple-encoders with these speaker tokens.
\fi

We employ three additional components: First, as a preliminary step, the model is pre-trained as a bi-encoder, i.e., standard CCL. We found this to improve results slightly (Appendix~\ref{app:ablations}). Second, we employ the same auxiliary NLI learning objective as is used in standard CCL.
Third, like in CCL, we indicate odd and even turns through additional speaker tokens, following~\citep{erker-etal-2023-imagination}.

\section{Application of Curved Contrastive Learning}\label{sec:problem}
This section describes how the bi-encoder and triple-encoder are used after training to solve the two previously introduced tasks~\citep{erker-etal-2023-imagination} \emph{sequence modeling} (Section~\ref{sec:seq}) and \emph{short-term planning} (Section~\ref{sec:planning}). Note, when using the triple-encoder as bi-encoder, we use the same setup as CCL bi-encoders.

\subsection{Dialog Sequence Modeling}\label{sec:seq}
In sequence modeling, given a context prefix $C = u_1, \dots, u_k$ from a dialog $u_1, \dots, u_n$, the task is to find the true utterance $u_{k + 1}$ among a set of randomly sampled future utterances. For every candidate utterance $u_f \in U_F$, the relative likelihood $p(u_f | C)$ (similarity score) is computed. For evaluation, we measure the rank of $p(u_{k + 1} | C)$ of all utterances in the test corpus at the same depth. The bi-encoder and triple-encoder differ in how $p(u | C)$ is computed.

\subsubsection{Bi-Encoder}\label{sec:seqBI}
Following \citet{erker-etal-2023-imagination}, with the bi-encoder the relative likelihood is computed as the cosine similarity between the candidate utterance and every context utterance (encoded separately) as 
% \begin{figure}[ht]
 %\vspace{5pt}
%\resizebox{0.5\textwidth}{!}{
\begin{equation*}
    p(u_f | C) = \sum\limits_{u_{i} \in C} cos(\enc([B] u_i), \enc([A] u_f)).
\end{equation*}
%}
%\vspace{5pt}
%\end{figure}

While the accumulation is very efficient and worked fairly well, we demonstrate that our triple encoder trained with C3L significantly improves the performance thanks to the extra contextualization.

%fairly well in chit-chat datasets like DailyDialog \citep{lin2020dialog}, it does not generalize to task-oriented dialogs like MDC \citep{li2018microsoft} due to the lack of contextualization as shown by the example in Figure \ref{fig:difficult}. In the following section we describe how we adapt the training objective of C3L to sequence modeling. 

\subsubsection{Triple-Encoder}
\begin{table}[]
    \centering
    \small
\begin{tabular}{c|ccccc|c|c}
& \multicolumn{5}{c}{B1} & Relative & Total \\
& & & & & & Growth & States \\
\hline
B2 & \( u_1 \) & \( u_2 \) & \( u_3 \) & \( u_4 \) & \( u_5 \) & & \\
\hline
\( u_1 \) & (X) & & & & & 0 & 0 \\
\( u_2 \) & 1 & (X) & & & & 1 & 1 \\
\( u_3 \) & 2 & 2 & (X) & & & 2 & 3 \\
\( u_4 \) & 3 & 3 & 3 & (X) & & 3 & 6 \\
\( u_5 \) & 4 & 4 & 4 & 4 & (X) & 4 & 10 \\
\end{tabular}
    \caption{Shown are the mean operations between utterances for a sequence length $n=5$. The cell values between [B1] and [B2] indicate the turn in which the relative state is computed. In contrast to the quadratic complexity of ConveRT ~\citep{henderson-etal-2020-convert}, our model lies within a linear complexity as shown by the number of relative growth. We provide a detailed complexity comparison in  Table \ref{tab:complex} in the appendix.}
    \label{tab:state}
\end{table}
Similar to the bi-encoder we accumulate the likelihood of a sequence based on the pairwise mixed representations of the entire sequence length $n$, i.e. the training window size $w$ does not apply during inference. We construct the relative likelihood for every candidate utterance $u_f \in U_F$ for a context $C := [u_1, .., u_n]$ of length $n$ as:
 %\begin{figure}[H]
 %\vspace{5pt}
%\resizebox{0.5\textwidth}{!}{
%\begin{equation}\label{eq:totalstate}
%    P(u_f | \text{C}) = \sum\limits_{i=1}^{n-1} \sum\limits_{j=i+1}^{n} \cos\left(\frac{\enc([B1]u_i) + \enc([B2]u_j)}{2}, \enc([A]u_f) \right)
%\end{equation}
%}
%\vspace{5pt}
%\end{figure}

\begin{equation}\label{eq:totalstate}
\small
    P(u_f | \text{C}) = \sum\limits_{i=1}^{n-1} \sum\limits_{j=i+1}^{n} \cos\left(\frac{\enc([B1]u_i) + \enc([B2]u_j)}{2}, \enc([A]u_f) \right)
\end{equation}
As the distributed pairwise mixed representations are superimposed (Equation \ref{eq:totalstate}), our embedding space emerges multiple local maxima in the latent space that enables a richer interaction for the candidate (\textbf{after space}) as shown in Figure \ref{fig:intro}. This stands in stark contrast to traditional search with only one context vector (like ConveRT \citep{henderson-etal-2020-convert}) where the candidate space is mapped to this one global maximum. This late interaction lets us build upon previous token-based techniques like ColBERT \citep{10.1145/3397271.3401075}. While Equation \ref{eq:totalstate} effectively captures their averaging approach, in Appendix \ref{sec:max} we also experiment with their default maximum-based approach, which is 100 times slower than simple averaging in our experimental context.

With our contextualized representations, the number of representations grows to the triangular numbers ($\frac{n(n + 1)}{2}$) as shown in the arising triangle of Table \ref{tab:state} (in the appendix). However, the relative number of computations at each turn is strictly \textbf{linear} (shown by the relative growth). Therefore at each turn, we simply compute all additional states from $t-1$ to $t$ as one matrix multiplication of size \((|U_F|, e\_dim)\) and \((n, e\_dim)^T\), where \(e\_dim\) denotes the embedding dimension and $|U_F|$ the size of the candidate space. Lastly, we add the score matrix of step $n-1$ and rank the candidates in the next step.

As the model is trained only on a fixed window size of $w$, taking the full triangle on longer sequence lengths might lead to unwanted distortions as distances out of this window are not part of the learning objective. Therefore, we also experiment with the $l$ last rows of Table \ref{tab:state}. In other words, we here only take the last $l$ utterances in the \texttt{[B2]} space and contextualize them with the entire sequence in the \texttt{[B1]} space with the previously discussed order constraint $[B1] u_i \And [B2] u_j \Leftrightarrow  i<j$. 
%Note that, at inference time using the $k$ last rows is 
Notably this $l$ last rows version with linear complexity has the same efficiency as the entire average since we only compute the latest row at each turn (equivalent to $l=1$ as shown in Table \ref{tab:state}).

\subsubsection{Efficiency Compared to CCL}
When it comes to the number of transformer computations triple-encoders enable a similar relativistic state accumulation in sequence modeling as the traditional CCL. As Figure~\ref{fig:relativeTimeDimension} demonstrates, during inference it is only necessary to encode the utterance at turn $n$ with the \texttt{[B2]} token and average it with every previous utterance $[[B1]u_1, \ldots, [B1]u_{n-1}]$. Only in the next turn it is necessary to encode the new utterance with \texttt{[B1]} which can be done during the dialog partner's turn. We provide a detailed complexity comparison between all approaches in Table \ref{tab:complex} in the appendix.

%\subsection{Pre-Filtering}\label{sec:prefiltering} %\todo{make an experiment to identify if an utterance is breaking linearity via anticipation}
%For retrieval, the size of the index of candidate representations is a crucial bottleneck in the speed of search. 
%While traditional utterance (re-)ranking methods like ConveRT \citep{henderson-etal-2020-convert} are trained to only predict the direct next utterance, curved contrastive learning (also in its contextual version) learns a distance measure that lets it retrieve representations over longer distances. Combined with its relative state accumulation, we can use the idle time while the dialog partner is speaking at the previous state $t-1$ to pre-filter future representations for step $t+1$. To our best knowledge, this task is unexplored and could potentially decrease the index size at step $t$ and therefore increase search speed significantly. We evaluate this pre-filtering with the same setup as for sequence modeling, with the only exception that the next utterance that we rank is not one but two turns away. 
%\todo{add one experiment showing that we can realize when the linearity breaks}

\subsection{Short-Term Planning}\label{sec:planning}
\citet{erker-etal-2023-imagination} have shown that the sequential information of CCL is especially useful to determine whether a candidate utterance is leading to a goal over multiple turns by just measuring their relative distance.  
The short-term planning experiments are conducted as follows: A dialog context $c[:l]$ of fixed length $l$ is given to a dialog transformer, which generates $100$ utterances for each context. The true utterance of dialog at that position is added to these $100$ candidates. 
We then rank all candidates by cosine similarity between the candidate (in the \textbf{before} space) and the goal $g$ (in the \textbf{after} space). This goal $g = c[l + g_d]$ is defined as the utterance of the true dialog $g_d$ turns in the future. 

\subsubsection{Bi-Encoder}
Here the true utterance should be closest to the goal in the imaginary space. We measure the cosine similarity between every candidate (in the \textbf{before} space) and the goal (in the \textbf{after} space) as $\forall c \in \text{Candidates}: cos(\enc([B] c,) \enc[A] g))$ where we rank the score of the true utterance. Notably, as mentioned by \citep{erker-etal-2023-imagination} the goals are picked at fixed but arbitrary positions. Hence, we can not ensure low ambiguity: For example, a response like "ok, okay" as the goal is achievable through various dialog paths, making 100\% accuracy unrealistic.

\subsubsection{Triple-Encoder}\label{sec:ltpTheory}
Through the relativistic property and the independence assumption in classical imaginary embeddings, the candidates are in no interaction with the context. With the triple-encoders, this shortcoming can be surpassed (1) through contextual aware training and (2) through contextual combination at inference. In particular, instead of determining the likelihood of candidates leading us to the goal over multiple turns as simple cosine similarity between the candidate and the goal, we combine the likelihood of the goal independently with its contextualized version. In particular, by the mean of the candidate $[B2]c$ with every context utterance $[[B1] u_1, \ldots, [B1] u_n]$ as the linear combination. The relative likelihood for a candidate $c$ to the other candidates is summarized as 
%\newline 
%\resizebox{0.5\textwidth}{!}{
%\begin{equation*}
%    cos(\enc([B2]c), \enc([A]g)) + \frac{1}{n} \sum\limits_{i=1}^{n} cos\left( \frac{\enc([B1]u_i) + \enc([B2]c)}{2}, \enc([A]g) \right)
%\end{equation*}
%}
\begin{align}
\begin{split}
     cos(\enc([B2]c), \enc([A]g)) +  \\ \frac{1}{n} \sum\limits_{i=1}^{n} cos\left( \frac{\enc([B1]u_i) + \enc([B2]c)}{2}, \enc([A]g) \right)
     \end{split}
\end{align}\newline

Here $n$ is the entire context length. We then rank the true utterance among the candidates.

\section{Experiments}\label{sec:experiments}
%\todo{jedes experiment beschreib effect auf einzelne RQ}
Our experiments are conducted on the newly introduced GTE (general-purpose text embedding) model  \citep{li2023general} as well as the RoBERTa-base models \cite{liu2019roberta} from the CCL paper \citep{erker-etal-2023-imagination} which we will use as baselines. Furthermore, we add a non-relativistic approach for sequence modeling evaluation, ConveRT \citep{henderson-etal-2020-convert}. Apart from that, we investigate ablations to our introduced triple-encoders using the same special tokens but without the curved property of the temporal dimension akin to curved contrastive learning as well as every component separately. 
The sequence modeling models are trained and evaluated on two datasets, DailyDialog \citep{li-etal-2017-dailydialog} and MDC \citep{li2018microsoft}, a task-oriented dialogue dataset. The models are also evaluated on zero-shot performance on PersonaChat \citep{zhang-etal-2018-personalizing}. We will furthermore evaluate the triple-encoder on short-term planning proposed by \cite{erker-etal-2023-imagination} where we evaluate our approach also with \texttt{Hits@k}. To compare to previous work we use the same setup, with candidates generated with DialoGPT \citep{zhang-etal-2020-dialogpt} (top\_p = 0.8, temperature=0.8) on history lengths of $2,5,10$ with goal distances of $1,2,3,4$. Our code including all hyperparameters can be found in our \href{https://github.com/UKPLab/acl2024-triple-encoders}{GitHub repository}. 
%, which uses, different from the paper~\citep{erker-etal-2023-imagination}, the more efficient dot-product over cosine similarity. 
%Upon acceptance, we will add our evaluation scripts to this package and release the triple-encoder loss within sentence transformers\footnote{\url{https://github.com/UKPLab/sentence-transformers}} \citep{reimers-gurevych-2019-sentence}.

\subsection{Self-Supervised Training}
All models are finetuned versions of the state-of-the-art text embedder GTE~\citep{li2023general}. We pre-train our models with CCL \citep{erker-etal-2023-imagination}, while we also experiment with training the triple-encoder from scratch. All models are trained on a window size of $w=5$, a batch size of $32$, a learning rate of $2\cdot10^{-5}$, a weight decay of $0.01$, utilize an Adam optimizer \citep{kingma2017adam} and use a linear warmup scheduler with $10\%$ of the training data as warmup steps. We perform model selection on the validation set after 10 epochs of training.

%  The models are performing best after $1-2$ epochs of training. 

\section{Evaluation \& Discussion}\label{sec:eval}
We start the evaluation with the sequence modeling performance of triple-encoder in Section \ref{sec:EvalSeq}, followed by the short-term planning performance in Section \ref{sec:EvalSTP}. We will address the research question \textbf{RQ1} by comparing triple-encoders trained with C3L to CCL bi-encoders. To answer \textbf{RQ2} we will furthermore compare to C3L bi-encoders (triple-encoder as bi-encoder), e.g training with C3L (Section~\ref{sec:triple}) but using the model at inference as bi-encoder by only using the \texttt{[B2]} token (Section~\ref{sec:seqBI}). We provide a comprehensive analysis on different training setups in the Appendix \ref{app:ablations} and an analysis of the contribution of every component in the triple-encoder setup during inference in sequence modelling (Appendix \ref{app:component}). In summary, we find that pre-training with CCL (which includes directional negatives) and then continuing training with triples yields the best performance. Furthermore, all our model components bring a benefit. 
\subsection{Sequence Modeling}\label{sec:EvalSeq}
%Sequence modelling is evaluated on one open dialogue corpus DailyDialog \citep{li-etal-2017-dailydialog} and one task-oriented corpus MDC \citep{li2018microsoft}.

\subsubsection{DailyDialog}
\begin{figure*}[htbp]
    \centering
    \includegraphics[width=1.0\textwidth]{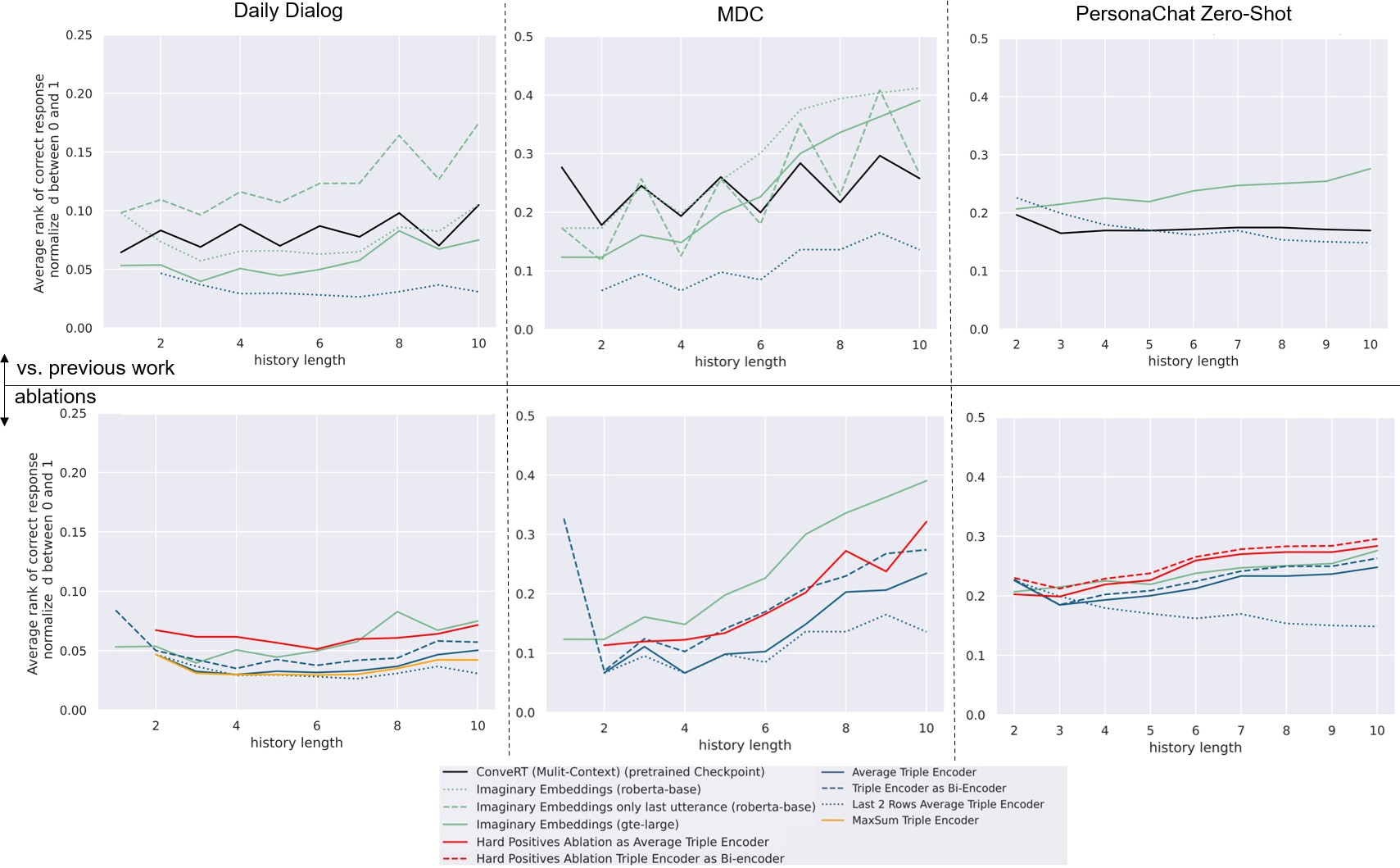}
    \caption{Sequence modeling performance via average rank ($\downarrow$) of true vs all utterances of the test set.}
    %We compare our triple-encoders (including ablations) to Imaginary Embeddings \citep{erker-etal-2023-imagination} and ConveRT \cite{henderson-etal-2020-convert}.
    % TODO: replace utterance $n$ with target utterance $E_T$
    \label{fig:DDSeq}
\end{figure*}

%\begin{figure}[H]
%    \centering
%    \includegraphics[width=0.5\textwidth]{images/seq_modeling.png}
%    \caption{Sequence Modeling Performance of triple-encoders vs Imaginary Embeddings \citep{erker-etal-2023-imagination}}
%    % TODO: replace utterance $n$ with target utterance $E_T$
%    \label{fig:DDSeq}
%\end{figure}

\begin{figure}
    \centering
    \includegraphics[width=0.5\textwidth]{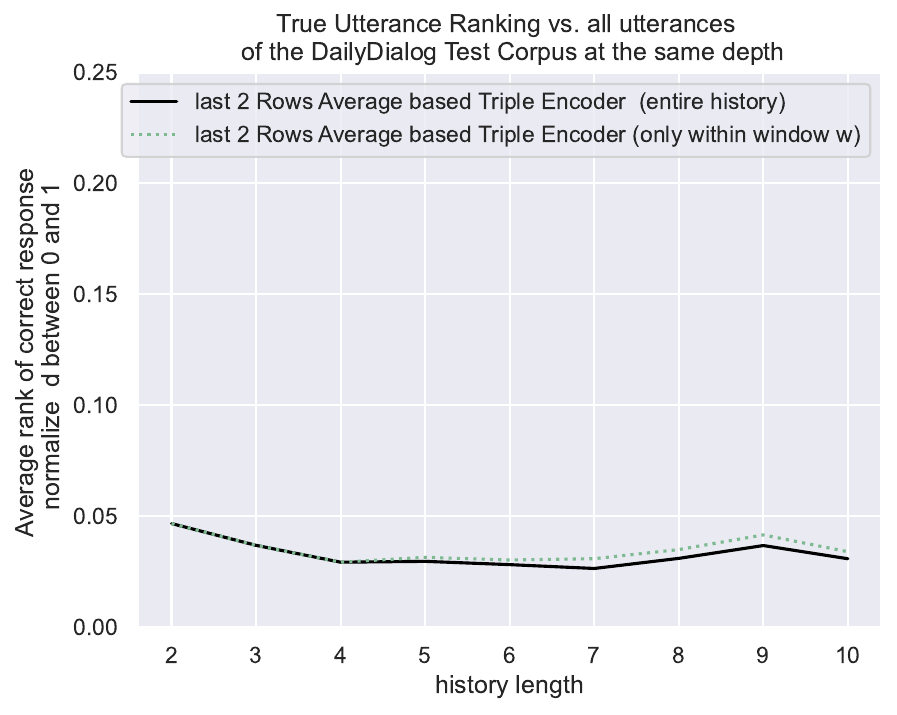}
    \caption{Shown are experiments on removing utterances outside the training window of $w=5$ compared to our version where we include outside-of-window utterances on the sequence modeling task (average rank $\downarrow$). We find on our most capable model, the Average Triple Encoder with $l=2$ (last 2 rows), that incorporating representations outside of the training window $w=5$ gains a 9.58\% lower average rank, demonstrating that our co-occurrence objective improves performance beyond its initial training window.}
    \label{fig:modelingWindow}
\end{figure}

\begin{table}[]
    \centering
    \small
    \begin{tabular}{cc}
    \hline
        $l$-last rows & Avg. Rank \\
        \hline
        l = 1 & 20.44 \\
        l = 2 & \textbf{19.75} \\
        l = 3 & 23.65\\
        l = 4 & 24.26\\
        \hline

    \end{tabular}
    \caption{$l$-last rows (of table \ref{tab:state}) Average triple-encoders performance in Sequence Modeling on DailyDialog. Here the l-last utterances in the \texttt{[B2]} space are contextualized with the entire sequence in the \texttt{[B1]} space.}
    \label{tab:k}
\end{table}
We compare the previously discussed architectures in terms of the sequence modeling performance on the DailyDialog corpus over different context lengths in Figure \ref{fig:DDSeq} (left). We start with the average triple-encoder that beats all baselines across all context lengths with an average rank of $21.25$, outperforming its non-curved (\textbf{hard positives ablation}) triple-encoder by $34.37\%$ and CCL (with the same GTE encoder base) (\textbf{RQ1}) by $31.46\%$. When it comes to the different variations, we find that MaxSim (average rank of $20.16$) on the entire triangle yields best performance within the context size of 5 (the training window). However, computing the maximum is 100 times slower, while performing only marginally better than averaging. Therefore, we recommend using average triple-encoders. Over context length of $5$ we find that the $l=2$ last rows variant of average triple-encoders performs best (Table~\ref{tab:k}). On this model we find in an ablation (Figure \ref{fig:modelingWindow}) that incorporating representations outside of the training window with size $w=5$ gains a $9.58\%$ lower average rank, demonstrating that our co-occurrence objective improves performance beyond its initial training window.

%Firstly, it is important to note how strong the new GTE-Large \citep{li2023general} model improved from the simple roberta-base \citep{liu2019roberta} model from \citep{erker-etal-2023-imagination} CCL by $28\%$. 

\subsubsection{Triple-Encoders as Bi-Encoders}
C3L demonstrate their versatility when used as bi-encoders \textbf{(RQ2)}. With the triple-encoder achieving an average rank of \(21.25\), and the GTE bi-encoder \citep{erker-etal-2023-imagination} achieving \(31.01\), the performance of the triple-encoder when treated as a bi-encoder sits impressively closer to the triple-encoder than to the bi-encoder with an average rank of \(25.48\). Our evaluations suggest that the difference is not merely attributed to the negatives. An ablation study using bi-positives and triple negatives achieves an average rank of \(27.30\), indicating that the positives play a pivotal role in narrowing the gap to \(25.48\).
This hints to a principle from neuroscience: 
\begin{quote}
        Neurons that fire together, wire together. \citep{https://doi.org/10.1002/sce.37303405110}
\end{quote} 
In the context of our triple-encoder, the co-occurrence of context utterances during training (as they "fire" together) leads to stronger associations or "wiring" between them in the embedding space, specifically by pushing co-occurring representations that wire together to a representation in the \textbf{after space} closer together. This leads to the phenomenon that even when processed separately, the embeddings have a stronger linear additivity to the candidate (after space) when being superimposed. We investigate this in more detail in Appendix \ref{app:random}. While training with triplets provides the model with rich contextual information, the persistence of learned associations during bi-encoder inference allows the model higher efficiency than triple-encoders with contextualization. 

\subsubsection{Task-Oriented Dialog Performance}
%\begin{figure}[H]
%    \centering
%    \includegraphics[width=0.5\textwidth]{images/seq_modelingMDC.png}
%    \caption{triple-encoder on MDC (task-oriented corpus)}
%    % TODO: replace utterance $n$ with target utterance $E_T$
%    \label{fig:TripleEncodersMDC}
%\end{figure}
One major shortcoming of CCL is its weak performance on task-oriented dialog corpora \citep{erker-etal-2023-imagination}. As shown in Figure \ref{fig:DDSeq} (middle), our curved triple-encoder improves upon the curved bi-encoder by $46\%$ significantly (\textbf{RQ1}). Overall we observe that contextualization brings the biggest benefit to task-oriented corpora, as both the non-curved and curved triple-encoders outperform bi-encoders. In contrast to \citet{erker-etal-2023-imagination}, we find that the curvature of triple-encoders is essential on task-oriented corpora as well, yielding a $20\%$ performance boost over the hard positives triple-encoders ablation. As Figure~\ref{fig:DDSeq} (middle) shows, on larger context size the $l=2$ triple-encoder outperforms the standard average triple-encoder similar to the DailyDialog experiments. Again, we observe that triple-encoder as bi-encoder also outperforms CCL (\textbf{RQ2}) substantially.

\subsubsection{Zero-Shot Performance}
%\begin{figure}[H]
%    \centering
%    \includegraphics[width=0.5\textwidth]{images/seq_modelingZero.png}
%    \caption{Zero-Shot Model Performance Comparison on PersonaChat}
%    % TODO: replace utterance $n$ with target utterance $E_T$
%    \label{fig:zeroshot}
%\end{figure}
For out-of-distribution dialogs on PersonaChat \citep{zhang-etal-2018-personalizing} in Figure \ref{fig:DDSeq} (right) we find that the 2-last rows of our emerging triangle of mixed representations (Table \ref{tab:state}) are crucial for generalization to larger context sizes. As the gap in the finetuned experiments is much smaller, we find that longer turn distances are much weaker modeled in zero-shot settings. Nonetheless, the hard positives ablation still performs significantly worse than using curved scores. The fact that our last 2 rows ($l=2$) average triple-encoder outperforms ConveRT shows that the co-occurrence objective has nonetheless strong generalization capability. Here we also observe how our distributed representations over ConveRT's one context vector information bottleneck comes into play. Initially, ConveRT demonstrates superior performance, but as shown in Figure~\ref{fig:DDSeq} (right), our model progressively improves with longer context lengths. It starts outperforming ConveRT when the context length reaches $5$ and continues to exhibit improvement, in contrast to ConveRT's performance plateau. Notably, the triple-encoder as bi-encoder generalizes also better on zero-shot scenarios compared to simple CCL \citep{erker-etal-2023-imagination} (\textbf{RQ2}).

\subsection{Short-Term Planning}\label{sec:EvalSTP}
\begin{table}[!htbp]
\centering
\small

\begin{tabular}{lccc}
\hline
Metric & Bi-Encoder & Triple-Encoder & Triple-Bi- \\
       & (CCL) & (ours) & Encoder (ours) \\
\hline
Hits@5 & 25.50 & 39.37 &  38.45 \\
Hits@10 & 34.99 & 48.44 & 46.82 \\
Hits@25 & 52.36 & 63.84 & 62.82 \\
Hits@50 & 71.73 & 79.17 & 78.63 \\
\hline
\end{tabular}
\caption{Average Hits@K Metrics in Short-Term Planning for our model and CCL \citep{erker-etal-2023-imagination}} 
% on history lengths of $2,5,10$ with goal distances of $1,2,3,4$ (except for history length of $10$ where due to low number of samples we limit the goal distances to $1$ and $2$.)
\label{tab:STPEval}
\end{table}
We evaluate the triple-encoder also in the short-term planning scenario on DailyDialog. As expected the extra contextualization helps on this task as shown in Table \ref{tab:STPEval}, most significantly on the \textbf{Hits@5} metric. We find that the gap from contextualized triple-encoder to the triple-encoder as bi-encoder is significantly closer than in other tasks. This demonstrates the versatility of the pre-training alone \textbf{(RQ2)} while the contextualization at inference shows additional small gains \textbf{(RQ1)}.

\section{Conclusion}
In this paper, we presented a novel approach for conversational sequence modeling, addressing the limitations of traditional methods such as ConveRT. Our triple-encoder leverages the concept of Curved Contrastive Learning and enhances it by incorporating contextualization through a Hebbian-inspired co-occurrence learning where representations that fire in a sequence together, wire together. This enables a more efficient and effective representation of dialog sequences without the need for additional weights, merely through local interactions, a first-of-its-kind approach that exhibits these self-organizing properties. As a result, our method outperforms single vector representation models on long sequences in zero-shot settings.

%Our experimental results demonstrate that the triple-encoders' contextualization model significantly outperforms existing models in various tasks \citep{RQ2}. Improving the average rank in sequence modeling on open-dialog by $36\%$ and task-oriented by $46\%$ compared to traditional Curved Contrastive Learning. Other than its processor, it could outperform ConveRT on zero-shot. Thanks to its distributed representations, particularly on longer distances. While we explored various late interaction methods, we find that the very efficient $k$-last rows average, with a dot product of length $2n$ ($n$ as sequence length), finds the best performance on long context sizes.

%Most interesting however is the performance of triple-encoders as bi-encoders during inference, outperforming all traditional CCL bi-encoders across all tasks \textbf{RQ2}. Most significantly on short term-planning, almost matching the performance of contextualized triple-encoders.

Our work demonstrates the distributed modularity of sequential representations by only mapping sequential properties within latent sub-spaces, i.e. all information is stored in the geometry of the latent space. For future work, we envision the exploration of triple-encoders for sequence modeling tasks other than dialog and story modeling.  To encourage the community to contribute in this direction we release our model and open-source our code.

%This could open new avenues in enhancing the sequential modularity, efficiency and contextual richness of  representations in dialogue systems and sequence modeling generally.

\newpage

\section{Limitations}
Building on \citet{erker-etal-2023-imagination} work we face similar limitations. In particular we address in this section the random splitting of our dataset in short-term planning, the use of synthetic data from LLMs to generate candidates replies, the generalizability to other datasets/tasks and response selection in the era of LLMs.

\textbf{Splitting data for the short-term planning experiments:} Like in \citet{erker-etal-2023-imagination}, our short-term planning results are limited by the fact that we split at fixed positions in the dialog, which might not necessarily be planable. While this suggests that the models perform slightly better if planning were always possible, it offers an unbiased comparison between the different models. 

\textbf{Usage of synthetic data in short-term planning experiments:} 
Additionally, the candidates for this task are generated by a large language model (LLM) where two issues can arise: (1) An utterance might lead to a goal that is not very likely given the context (see \citet{erker-etal-2023-imagination}) or (2) where the true utterance is out of distribution of the LLM candidates and this true utterance can only reach the goal. 

\textbf{Datasets:}
One further limitation of our work is that our models are only tested on three dialog datasets and only one story generation dataset. 

\textbf{Response selection in era of LLMs:}
While Large Language models are becoming more and more popular in response generation, they still suffer from hallucinations \citep{bouyamourn-2023-llms}, which is why retrieval is still popular, especially in legal and medical domains \citep{louis2022statutory, shi-etal-2023-midmed}. 
% scheiß modelle 
% fast nur dialog und text gerneration 
% eigentlich kaum retrieval based --> sequential modularity sequezen abbilden für future work  
% 
\section{Ethics}
Like other work \citep{Schramowski2022, prakash-lee-2023-layered}, our models can have induced biases based on their training data. While we do not adress the concerns in this paper, all datasets that are used in our experiments are publicly available and do not include any sensitive information to the best of our knowledge.
%In the past the environmental cost of machine learning models has been a subject of controversy \citep{strubell-etal-2019-energy}. However, in contrast to conventional approaches, our approach is more efficient as each utterance in a sequence can be encoded independently. 
% green NLP, less compute 
% all NLP systems biases from underlying data

\section{Acknowledgement}
This research work has been funded by the German Federal Ministry of Education and Research and the Hessian Ministry of Higher Education, Research, Science and the Arts within their joint support of the National Research Center for Applied Cybersecurity ATHENE. Furthermore, this work has been funded by the LOEWE Distinguished Chair “Ubiquitous Knowledge Processing”, LOEWE initiative, Hesse, Germany (Grant Number: LOEWE/4a//519/05/00.002(0002)/81).

\bibliography{anthology,custom}
\bibliographystyle{acl_natbib}

\appendix

\section{Maximum Similarity}\label{sec:max}
\begin{algorithm}
\small
\begin{algorithmic}
\Require $bmm\_result$, $index\_tuples$

\For{$sample \leftarrow 0$ to $bmm\_result.shape[0]-1$}
    \For{$u_f \leftarrow 0$ to $bmm\_result.shape[1]-1$}
        \State $utt\_used \leftarrow \text{empty set}$
        \State $bmm\_sample \leftarrow bmm\_result[sample][u_f]$
        \State $i\_sort \leftarrow \text{argsort}(bmm\_sample, \text{desc})$
        \State $tuples\_sort \leftarrow [index\_tuples[i] \text{ for } i \text{ in } i\_sort]$

        \State $sum \leftarrow 0$, $counter \leftarrow 0$
        
        \For{$i$ in $i\_sort$}
            \If{$tuples\_sorted[i][0] \not\in utt\_used$ \textbf{or} 
            \Statex\hspace{\algorithmicindent}\hspace{\algorithmicindent}\hspace{\algorithmicindent}
                $tuples\_sorted[i][1] \not\in utt\_used$}
            %\If{$tuples\_sortedles[i][0]$ $\not\in$ $tuples\_used$ and $tuples\_sortedles[i][1]$  $\not\in$  $tuples\_used$}
                \State $sum \leftarrow sum + bmm\_sample[i]$
                \State $counter \leftarrow counter + 1$
                \State add $tuples\_sorted[i][0]$ to $utt\_used$
                \State add $tuples\_sorted[i][1]$ to $utt\_used$
            \EndIf
        \EndFor
        \State $bmm\_result[sample][u_f] \leftarrow sum/ counter$
    \EndFor
\EndFor

\end{algorithmic}

\caption{MaxSim for triple-encoders}\label{algorithm:maxsim}
\end{algorithm}
In the maximum similarity-based approach we compute the batched matrix multiplication (BMM) as in the average similarity based version. Notably, our MaxSum Algorithm \ref{algorithm:maxsim} expects the entire state (entire triangle), which can be concatenated with the BMM matrices from previous turns. For each candidate-context pair we sort the scores of every pairwise contextualized representations in decreasing order. Similar to query representations of ColBERT, we then we add every score only if any of the utterances in the tuples was not yet part of the sum. In contrast to the simple average, the number of states can differ from candidate to candidate. Therefore, we have to average the result by dividing by the number of states.

\section{Ablation Studies}\label{app:ablations}

\begin{table}[]
\center
\small 
\begin{tabular}{|c|c|c|}
\hline
Ablation Analysis                                               & \begin{tabular}[c]{@{}c@{}} pre-trained  \\ with CCL \end{tabular}                           & \begin{tabular}[c]{@{}c@{}}Test \\ (avg. rank)\end{tabular} \\ \hline
triple-encoder   & yes             & 21.25                                                                      \\ \hline
triple-encoder   & no                     & 23.02                                                                      \\ \hline
\begin{tabular}[c]{@{}c@{}}triple-encoder\\ (directional negatives)\end{tabular}  & yes          & 25.68                                                                      \\ \hline
\begin{tabular}[c]{@{}c@{}}triple-encoder as bi\\ (bi pos + triple negatives)\end{tabular} & yes  & 27.30                                                                      \\ \hline
CCL GTE ablation  &  only  & 31.01                                                                     \\ \hline
\begin{tabular}[c]{@{}c@{}}triple-encoder \\ (hard positives / no curvature)\end{tabular} & no  & 32.39                                                                      \\ \hline
\end{tabular}
\caption{Ablation Analysis of triple-encoders. The first model utilizes a pre-trained checkpoint from curved contrastive learning (CCL)(Note that CCL has directional negatives), the second model is trained from scratch, and the third one utilizes directional negatives. Following is an ablation that uses bi-positives and triple negatives to show the improvement not only comes from the harder negatives. The last and worst ablation is a triple-encoder that is given hard positives instead of the relativistic distance curvature, the essence of Curved Contrastive Learning.}
\label{tab:tripleAblation}
\end{table}
Looking at the ablation study Table \ref{tab:tripleAblation}, we observe that pre-training a triple-encoder with (bi) curved contrastive learning (which has directional negatives) and then continuing with triplet loss (without directional negatives) yields the best performance. Followed by the triplet encoder trained from scratch and the triple-encoder with directional negatives. While all triple-encoders with the turn distance curvature (essence of CCL) yield better performance than bi-encoders, the ablation of utilizing triplet negatives but bi-positives already improves on simple CCL. Lastly, we compare our C3L triple-encoder to the triple-encoder without the curvature of scores, in other words, a triplet encoder only having hard positives. As the results show, it is $58.7\%$ worse than curved triple-encoders, showing the fundamental necessity of the temporal curvature of curved contrastive learning for sequence modeling on \textbf{relative}/modular components. 

\section{Component Analysis of Triple-Encoder}\label{app:component}
\begin{table*}[!htbp]
\centering
\begin{tabular}{|c|c|c|}
\hline
Description & Mathematical Definition                                                                 & \begin{tabular}[c]{@{}c@{}}Test \\ (average \\ rank)\end{tabular} \\ \hline
Triple-Encoder & $\sum\limits_{i=1}^{n-1} \sum\limits_{j=i+1}^{n} \cos^*\left( \frac{\enc([B1]u_i) + \enc([B2]u_j)}{2}, \enc([A]C_{n+1}) \right)$                                                                    & 21.25                                                                      \\ \hline
 \begin{tabular}[c]{@{}c@{}}Triple-Encoder\\ + bi-like [B2]\end{tabular} & $\begin{aligned}
&\sum\limits_{i=1}^{n-1} \sum\limits_{j=i+1}^{n} \cos^*\left( \frac{\enc([B1]u_i) + \enc([B2]u_j)}{2}, \enc([A]C_{n+1}) \right) \\
&+ \sum\limits_{h=1}^{n} \cos^*(\enc([B2]  u_h), \enc([A]  C_{n+1}))
\end{aligned}  $                                                             & 21.92                                                                      \\ \hline
direct neighbors & $\sum\limits_{i=1}^{n-1} \cos^*\left( \frac{\enc([B1]u_i) + \enc([B2]u_{i+1})}{2}, \enc([A]C_{n+1}) \right)$ & 24.88                                                                      \\ \hline
\begin{tabular}[c]{@{}c@{}} bi-like [B1] \\ and bi-like [B2]\end{tabular}  & $\sum\limits_{h=1}^{n} cos^*(\enc([B2]  u_h), \enc([A]  C_{n+1})) + cos^*(\enc([B1]  u_h), \enc([A]  C_{n+1}))$                                                                   & 25.08                                                                      \\ \hline
Mean with only [B2] & $\sum\limits_{i=1}^{n-1} \sum\limits_{j=i+1}^{n} \cos^*\left( \frac{\enc([B2]u_i) + \enc([B2]u_j)}{2}, \enc([A]C_{n+1)} \right)$                                                                       & 25.40                                                                      \\ \hline
bi-like [B2] &$\sum\limits_{h=1}^{n} cos^*(\enc([B2]  u_h), \enc([A]  C_{n+1}))$                                                                 & 25.48                                                                     \\ \hline

Mean with only [B1] &$\sum\limits_{i=1}^{n-1} \sum\limits_{j=i+1}^{n} \cos^*\left( \frac{\enc([B1]u_i) + \enc([B1]u_j)}{2}, \enc([A]C_{n+1}) \right)$                                                                        & 33.45                                                                     \\ \hline
\end{tabular}
\caption{Component analysis of one triple-encoder. The input variation is the simple triple-encoder as described in \ref{sec:triple}. As the mean operation might lose information of the true utterance, we add a version (second model) where we add the representations as a simple bi-encoder to the triplets. Following, we consider only direct neighbors where $[B1]u_i$ and $[B2] u_j$ only if $j - i = 1$. The following bi-like models are just bi-encoder versions of the triple-encoder while the mean with only  $[B1]$ or only $[B2]$ study the significance of using the distinct subspaces in the \textbf{before} space.}
\label{tab:tripleComponent}
\end{table*}
We continue with the component analysis on the best triple-encoder from Table \ref{tab:tripleComponent}. The normal input of the triple-encoder yields the best results. Since the mean operation of triple-encoders loses information of the original utterances, we added the normal bi-encoder cosine operation to the means, which reduced the performance. We note that the \texttt{[B1]} and \texttt{[B2]} tokens are essential, as the means between \texttt{[B1]} and \texttt{[B1]} as week as  \texttt{[B2]} and \texttt{[B2]} reduce the performance drastically. Interestingly, \texttt{[B2]} is significantly better than \texttt{[B1]} in both means with itself as well as alone as a bi-encoder. This makes sense as the utterance closer to the current turn should have a higher impact on a candidate's utterance than the ones further away. 
It's especially noteworthy that direct neighbor contextualization, which only accounts for adjacent utterance pairs, performs competitively compared to the combined bi-encodings of \texttt{[B1]} and \texttt{[B2]}. This underscores the value of non-local neighbor contextualization, which improves performance by 18\%.

%\section{speed comparison}
%already shown in the previous work by CCL. We do not count encoding the B1 representations as waiting time as it can be done in idle time while the dialog partner is speaking --> no effect on inference. show complexity of erker et al when it comes to the encoding

\section{Representations that Fire Together, Wire Together}\label{app:random}
\begin{figure*}
    \centering
    \includegraphics[width=\textwidth]{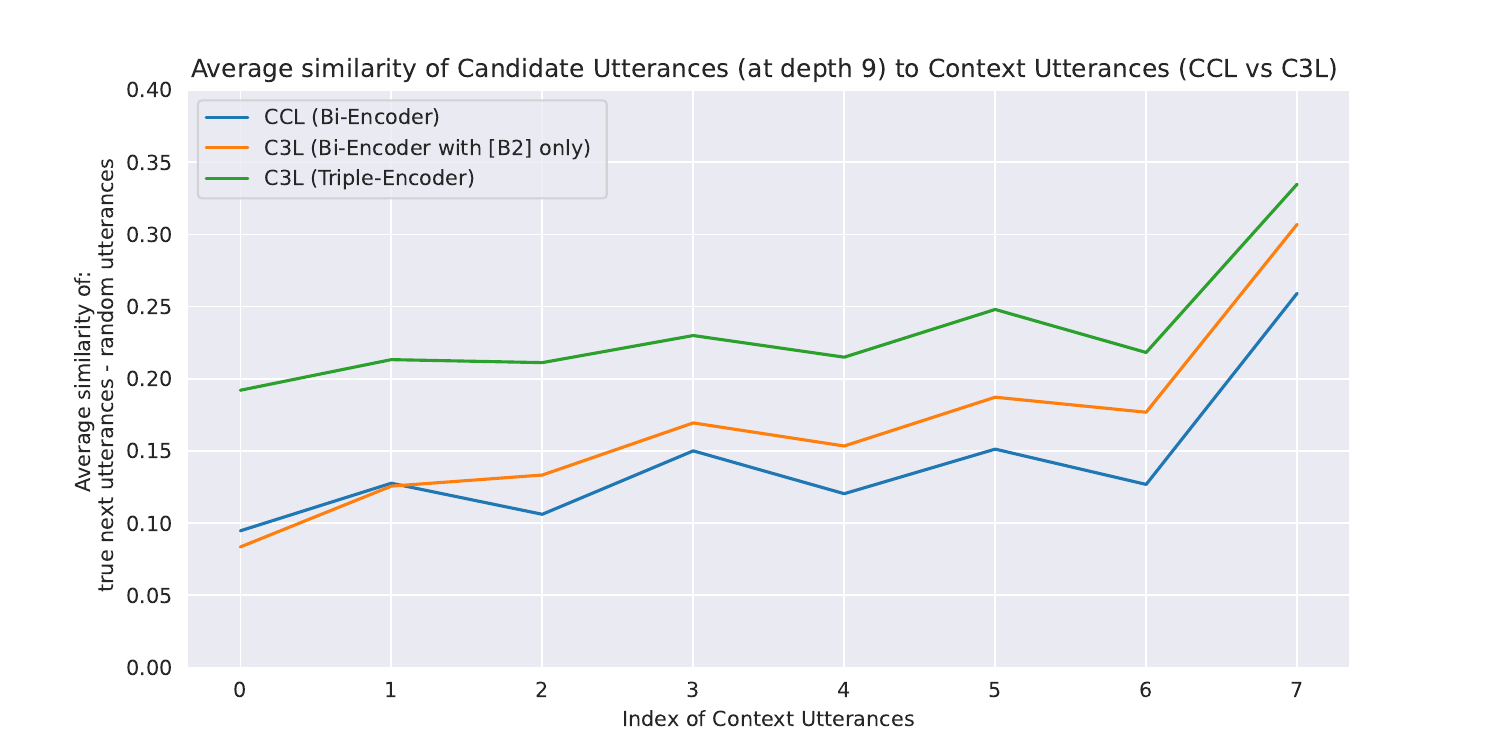}
    \caption{Additive properties of a true candidate utterance vs random utterances at turn 9 for every context utterance.}
    \label{fig:contextSim}
\end{figure*}

% \usepackage{multirow}
\begin{table*}[]
\begin{tabular}{|c|l|ll|l|}
\hline
\multirow{2}{*}{\begin{tabular}[c]{@{}c@{}}Inference \\ Type\end{tabular}} & \multicolumn{1}{c|}{\multirow{2}{*}{\begin{tabular}[c]{@{}c@{}}Approach with \\ Special Token\end{tabular}}} & \multicolumn{2}{c|}{Utterance in after space}                                                                                                                        & \multicolumn{1}{c|}{\multirow{2}{*}{\begin{tabular}[c]{@{}c@{}}Factor\\ Avg. sim correct -\\ Avg. sim random\end{tabular}}} \\ \cline{3-4}
                                                                           & \multicolumn{1}{c|}{}                                                                                        & \multicolumn{1}{l|}{\begin{tabular}[c]{@{}l@{}}Avg. Similarity \\ Random\end{tabular}} & \begin{tabular}[c]{@{}l@{}}Avg. Similarity\\ correct utterance\end{tabular} & \multicolumn{1}{c|}{}                                                                                                          \\ \hline
\multirow{2}{*}{Bi-Encoder}                                                & CCL ({[}BEFORE{]})                                                                                           & \multicolumn{1}{l|}{0.0659}                                                            & 0.2190                                                                      & 0.1531                                                                                                                         \\ \cline{2-5} 
                                                                           & C3L ({[}B2{]})                                                                                               & \multicolumn{1}{l|}{\textbf{-0.0031}}                                                  & 0.1616                                                                      & \textbf{0.1657}                                                                                                                \\ \hline
\multicolumn{1}{|l|}{Triple-Encoder}                                       & C3L ({[}B1{]} \& {[}B2{]})                                                                                   & \multicolumn{1}{l|}{0.0201}                                                            & 0.286                                                                       & \textbf{0.2659}                                                                                                                \\ \hline
\end{tabular}
\caption{Average similarity of bi-encoder sequences (before space)in C3L and CCL to the correct utterance vs random utterances (after space).}
\label{tab:randomcorrect}
\end{table*}
We start the investigation of stronger additive properties of C3L over CCL in the bi-encoder setup by comparing the average similarity of sequences to the correct utterance and random sampled utterances. We use the same setup as in the sequence modeling evaluation on the test set. While the absolute similarity of CCL to correct utterances in the bi-encoder setup is greater than in C3L, Table \ref{tab:randomcorrect} reveals that the similarity of C3L to random utterance is much closer to the target similarity of $0$ for hard negatives, demonstrating stronger discriminative properties. Specifically, we find that similarity difference from random utterances to correct utterances is greater in C3L compared to CCL. However, to demonstrate the stronger additive properties, a stronger contribution of each context utterance within sequences to candidate utterances has to be shown. Hence, we measure for every context utterance in all contexts of size $8$, the difference between the correct and the average random similarity. While for the bi-encoders each utterance is one representation, in the triple-encoder setup we have $n-1$ mixtures of each utterance which we aggregate (mean) for each utterance respectively. Our results in Figure \ref{fig:contextSim} show, that the additive properties over random utterances are significantly stronger over the entire history in C3L compared to CCL, thanks to our introduced co-occurrence learning objective. In general we observe that the latest utterance has the strongest contribution, while the influence of utterances from our dialog partner are in general more important shown by the fluctuation between odd and even turns. As each triple encoder utterance contains a mixture of all $n-1$ context utterances, further away utterances decay less strongly as in the bi-encoder setup. For bi-encoders, the gap between C3L \& CCL becomes closer over longer distances. While CCL looses a lot of information already after the last utterance, C3L bi-encoders have a much more steady decline as the information (wiring) of close utterances through its training objective is significantly better preserved.

%CCL however, drops at the beginning most abruptly and then decays just slightly while C3L as bi-encoder decays more steadily as someone would expect from the learning objective. The  

\begin{table*}[!htbp]
\centering
\begin{tabular}{lccc}
\hline
\textbf{Model} & \textbf{\begin{tabular}[c]{@{}c@{}}Number of utterance \\ encodings per step\end{tabular}} & \textbf{\begin{tabular}[c]{@{}c@{}}Number of cosine \\ similarities per step\end{tabular}} & \textbf{Performance} \\ \hline
CCL & \( \mathcal{O}(1) \) & \(\mathcal{O}(|C|) \) & + \\
C3L + Bi-encoder (Ours) & \( \mathcal{O}(1) \) & \( \mathcal{O}(|C|) \) & ++ \\
C3L + Triple-encoder (Ours) & \( \mathcal{O}(1) \) & \( \mathcal{O}(|C| \cdot n) \) & +++ \\
ConveRT & \( \mathcal{O}(n) \) & \( \mathcal{O}(|C|) \) & ++ \\ \hline
\end{tabular}
\caption{Computational complexity comparison between our Triple Encoder, CCL and ConveRT where $n$ represents the sequence length, $|C|$ the number of candidate utterances. Note that each encoding is very expensive compared to the efficient and highly parallelizable cosine similarity operations.}
\label{tab:complex}
\end{table*}
%\begin{table}[H]
%\centering
%\begin{tabular}{|c|c|c|c|c|c|c|c|c|}
%\hline
% & $u_1$ & $u_2$ & $u_3$ & $u_4$ & $u_5$ & $u_6$ & $u_7$ & $u_8$ \\
%\hline
%$u_1$ & (X) &   &   &   &   &   &   &   \\
%\hline
%$u_2$ & 0.17 & (X) &   &   &   &  &  &  \\
%\hline
%$u_3$ & 0.13 & 0.18 & (X) &  &  &  &  &  \\
%\hline
%$u_4$ & 0.21 & 0.21 & 0.24 & (X) &  &  &  &  \\
%\hline
%$u_5$ & 0.14 & 0.20 & 0.16 & 0.20 & (X) &  &  &  \\
%\hline
%$u_6$ & 0.23 & 0.22 & 0.26 & 0.23 & 0.27 & (X) &  &  \\
%\hline
%$u_7$ & 0.15 & 0.21 & 0.17 & 0.22 & 0.18 & 0.22 & (X) &  \\
%\hline
%$u_8$ & 0.32 & 0.31 & 0.35 & 0.31 & 0.37 & 0.32 & 0.38 & (X) \\
%\hline
%\end{tabular}
%\caption{Average Similarity of Triple Encoder to correct utterance %(depth 9) - random utterances for all dialogues of size 8}
%\end{table}

\section{Children Book Test}\label{sec:CBT}
Apart from dialog we also experiment with text generation within the Children Book Test dataset \cite{childrenbook}. 
\subsection{Setup}
The dataset is already split into a list of sentences for each story, which we treat similarly to utterances in our dialog setup. Apart from speaker tokens that are removed,  we apply our method in the same way as for dialogs. We train a simple bi-encoder with CCL \cite{erker-etal-2023-imagination}, triple encoders with C3L as well as its hard positive ablation. We evaluate the technique by ranking the next sentence.
\subsection{Evaluation}
Similar to our dialog results \ref{sec:eval}, we observe that Triple Encoders are improving significantly over CCL with an increase of $39.85\%$ in average rank (\textbf{RQ1}). This can also be observed for the triple as bi- encoder version that outperforms the hard positives ablation until a sequence length of $8$ where contextualization seems to become more important than the relative distance objective of C3L (\textbf{RQ2}). We explore different settings for last $l$ rows in Table \ref{tab:k-CBT}. We find $l=1$ performs best for sequences longer than 4 sentences. We believe that $l=2$ is worse here as the speaker tokens are absent and therefore taking two over longer distances might lead to distortions.

% simple next sentence ranking for text generation sequence modeling. use paper: \cite{a6d8b29b29664b0fb783844ec40ad66e}, created dataset sentences of "NE" - rank next

% todo all different k's

\begin{table}[]
    \centering
    \small
    \begin{tabular}{cc}
    \hline
        $l$-last rows & Avg. Rank \\
        \hline
        l = 1 & \textbf{170.57} \\
        l = 2 & 177.14 \\
        l = 3 & 192.44 \\
        l = 4 & 201.85 \\
        \hline

    \end{tabular}
    \caption{$l$-last rows average triple-encoders performance in sequence modeling on the Children Book Test. Notably, the average triple encoder achieves an average rank of $185.56$. Similarly to the dialog performance, it lies between $l=2$ and $l=3$}
    \label{tab:k-CBT}
\end{table}
\begin{figure*}[h]
    \centering
    \includegraphics[width=\textwidth]{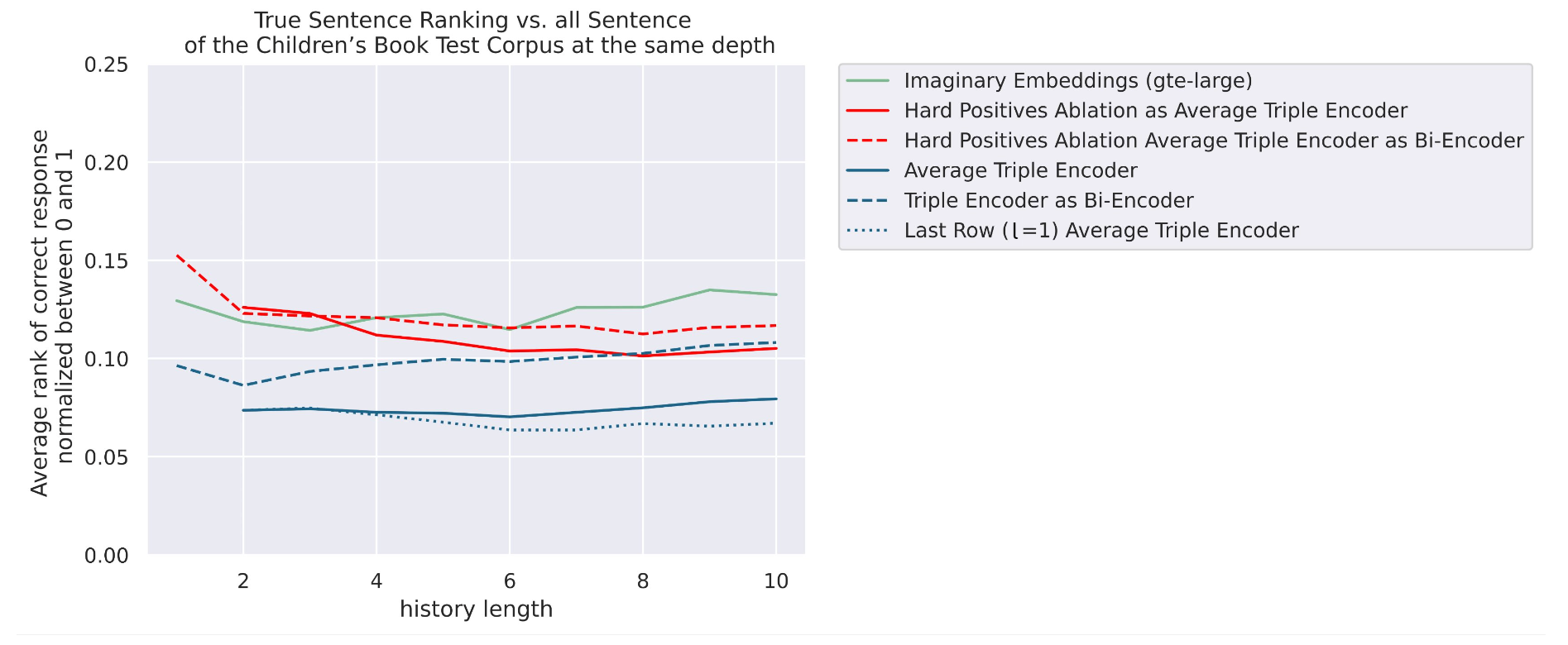}
    \caption{Sequence Modeling performance on next sentence prediction of Children Book Test corpus.}
    % TODO: replace utterance $n$ with target utterance $E_T$
    \label{fig:trajectory}
\end{figure*}

\end{document}